\documentclass[lettersize,journal]{IEEEtran}
\usepackage{amsmath,amsfonts}
\usepackage{array}
\usepackage[font=normalsize,labelfont=sf,textfont=sf]{subfig}
\usepackage{textcomp}
\usepackage{stfloats}
\usepackage{url}
\usepackage{verbatim}
\usepackage{cite}
\usepackage[linesnumbered,ruled,vlined]{algorithm2e}
\usepackage{algpseudocode}
\usepackage{siunitx}
\usepackage{xcolor}
\usepackage{multirow}
\usepackage{makecell}
\usepackage{colortbl}
\usepackage{xcolor}
\usepackage{tabularx}
\usepackage{arydshln} 
\usepackage{booktabs}
\usepackage{hyperref}
\usepackage{spreadtab}
\usepackage{graphicx}
\usepackage{caption}

\usepackage[margin=1in]{geometry}
\usepackage{titlesec}
\begin{document}

\title{Weakly Supervised Multimodal Temporal Forgery Localization via Multitask Learning}

\author{Wenbo Xu,
        Wei Lu, \IEEEmembership{Member,~IEEE},
        Xiangyang Luo
        
\thanks{This work is supported by the National Natural Science Foundation of China (No.62441237, No.62172435). ($Corresponding \ authors: Wei\ Lu$)}
\thanks{Wenbo Xu, Wei Lu are with the School of
Computer Science and Engineering, MoE Key Laboratory of Information Technology, Guangdong Province Key Laboratory of Information Security Technology, Sun Yat-sen
University, Guangzhou 510006, China (email: xuwb25@mail2.sysu.edu.cn; luwei3@mail.sysu.edu.cn).}
\thanks{Xiangyang Luo is with the State Key Laboratory of Mathematical Engineering and Advanced Computing, Zhengzhou 450002, China. (e-mail: luoxy$\_$ieu@sina.com)}
}



\maketitle
\begin{abstract}
The spread of Deepfake videos has caused a trust crisis and impaired social stability.
Although numerous approaches have been proposed to address the challenges of Deepfake detection and localization, there is still a lack of systematic research on the weakly supervised multimodal fine-grained temporal forgery localization (WS-MTFL).
In this paper, we propose a novel weakly supervised multimodal temporal forgery localization via multitask learning (WMMT), which addresses the WS-MTFL under the multitask learning paradigm. 
WMMT achieves multimodal fine-grained Deepfake detection and temporal partial forgery localization using merely video-level annotations.
Specifically, visual and audio modality detection are formulated as two binary classification tasks. 
The multitask learning paradigm is introduced to integrate these tasks into a multimodal task.
Furthermore, WMMT utilizes a Mixture-of-Experts structure to adaptively select appropriate features and localization head, achieving excellent flexibility and localization precision in WS-MTFL.
A feature enhancement module with temporal property preserving attention mechanism is proposed to identify the intra- and inter-modality feature deviation and construct comprehensive video features.
To further explore the temporal information for weakly supervised learning, an extensible deviation perceiving loss has been proposed, which aims to enlarge the deviation of adjacent segments of the forged samples and reduce the deviation of genuine samples.
Extensive experiments demonstrate the effectiveness of multitask learning for WS-MTFL, and the WMMT achieves comparable results to fully supervised approaches in several evaluation metrics.
\end{abstract}

\begin{IEEEkeywords}
Multimodal Deepfake detection, weakly supervised, multitask learning, temporal forgery localization.
\end{IEEEkeywords}

\section{Introduction}
\IEEEPARstart{G}{enerative} artificial intelligence has rapidly advanced in recent years, and existing artificial intelligence generative models could generate high-quality multimedia content such as image, audio, video, etc.
Deepfake, a typical application of generative models, allows for forging media content of actual people or generating fictional content.
Initially, Deepfake focused mainly on single-modal forgeries.
More recently, Deepfake has been utilized to create more complex and vivid multimodal forged videos, in both visual and audio modalities \cite{10168141,10024806,wu2025weakly}.
The misuse of Deepfake represents a substantial threat to individual privacy, copyright protection, and social stability.
Therefore, research on more general and robust multimodal deepfake detection paradigms is of critical academic value and practical significance.

\begin{figure}[!t]
\centering
\includegraphics[width=0.9\linewidth]{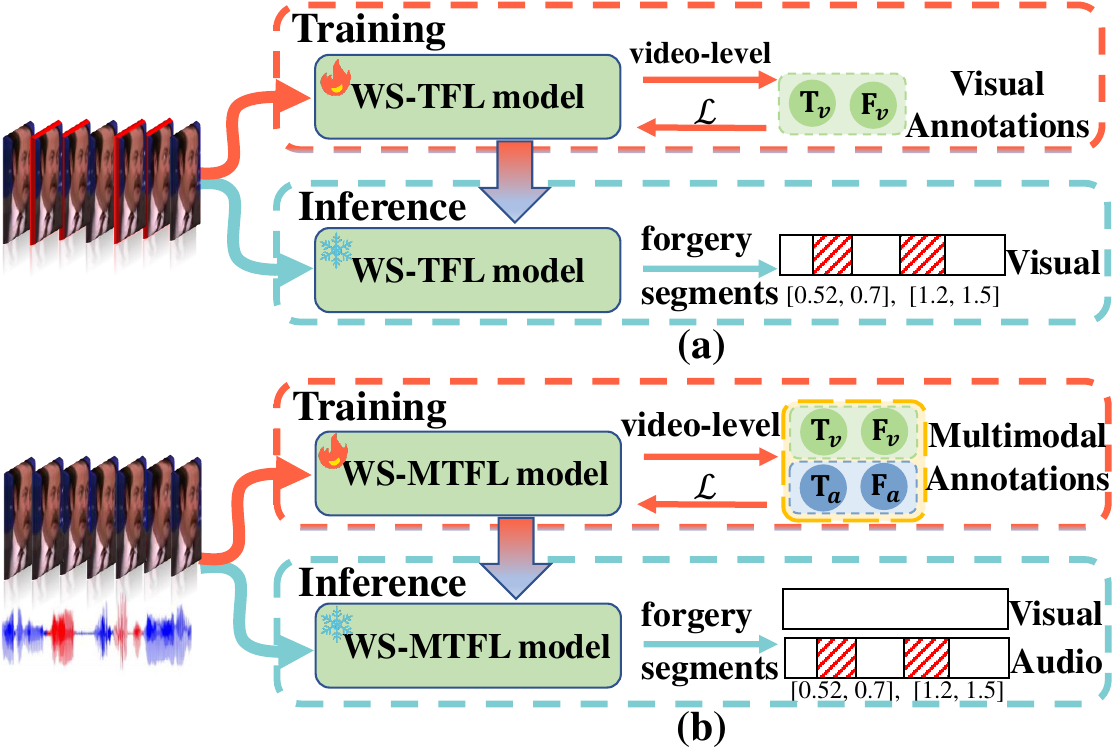}
\caption{Schematic of WS-TFL and WS-MTFL. 
(a) the model is trained with single-modal annotations and predicts forged segment timestamps during inference.
(b) trained with multimodal annotations, the model predicts both modality-specific forgery types (both genuine, both forged, visual forged only, audio forged only) and corresponding timestamps.}
\label{intro_ws_mtfl}
\end{figure}

There are basically three categories of Deepfake forensic tasks: classification, spatial forgery localization, and temporal forgery localization.
The classification tasks have drawn significant interest from researchers since the beginning of Deepfake forensics, such as single-modal binary classification \cite{10168141}, multimodal fine-grained classification \cite{yin2024fine}, and the generalizability of classification models \cite{10516609}.
The spatial forgery localization task primarily focuses on identifying the forged regions for Deepfake images \cite{10702428}.
As Deepfake technology evolves, only several segments in the video are forged (in visual or audio modality) in complex forgery scenarios.
To address the above-mentioned challenge, the temporal forgery localization task (TFL) is proposed to locate the start and end timestamps of forged segments \cite{chugh2020not}.
TFL has been explored in both fully supervised and weakly supervised paradigms.
Fully supervised TFL (FS-TFL) approaches typically rely on frame-level or timestamp annotations.
For instance, BA-TFD+ \cite{cai2023glitch} and AVTFD \cite{liu2023audio} attempted to achieve timestamp localization of forged segments using frame-level classification, while UMMAFormer \cite{zhang2023ummaformer} introduced a regression head to predict timestamps directly.
However, they require elaborate frame-level or timestamp annotations, which are usually costly and time-consuming.
To cope with the dilemma of FS-TFL, weakly supervised TFL (WS-TFL) has been proposed, which utilizes only video-level annotations for training and achieves timestamp localization during inference.
However, current WS-TFL approaches primarily focus on single-modal Deepfake, such as visual \cite{10.1109/TIFS.2025.3533906,10887844} or audio modality \cite{wu2025weakly}, and the multimodal temporal forgery localization is insufficiently explored.

In this paper, we propose weakly supervised multimodal fine-grained temporal forgery localization (WS-MTFL) task, which aims at modality-specific forgery type detection and the corresponding timestamp localization of forged segments.
The schematic diagrams of WS-TFL and WS-MTFL are shown in Fig. \ref{intro_ws_mtfl}.
The main challenges of WS-MTFL are: 1) the balance between multimodal interaction and single-modal forgery traces identification;
and 2) mine subtle temporal forgery traces for weakly supervised learning.
To address the above challenges, we propose weakly supervised multimodal temporal forgery localization via multitask learning (WMMT).
The multitask learning (MTL) paradigm is introduced to formulate the WS-MTFL into three related tasks: visual, audio, and multimodal task.
The MTL enables the WMMT to extract shared representations across tasks while retaining modality-specific features.
Besides, we design a modality-aware expert selection mechanism to adaptively select appropriate features and localization head during inference, thereby balancing multimodal interaction and single-modal forgery traces identification.
Moreover, a temporal property preserving attention mechanism (TPPA) is proposed to explore more forgery traces for multimodal temporal forgery localization.
Additionally, an extensible deviation perceiving loss is introduced to mine further temporal information for weakly supervised learning.

Specifically, the proposed WMMT contains four modules: feature extraction, feature enhancement, multitask loss, and temporal forgery localization.
The feature extraction module first extracts visual and audio features of a given video using pre-trained models, respectively.
Subsequently, the enhanced visual and audio features are obtained by the feature enhancement module, which aims to identify the intra- and inter-modality feature deviation.
During the training phase, the enhanced visual features, enhanced audio features, and multimodal (visual + audio) features are used to train three localization heads: the visual localization head, the audio localization head, and the multimodal localization head.
Each localization head is accompanied by a classification head for video-level modality-specific forgery type prediction.
In the MTL paradigm, the multitask loss jointly guides the training of three tasks together to improve the learning of each task.
Moreover, an extensible deviation perceiving loss is introduced for weakly supervised learning during training, aiming to enlarge the temporal deviation of forged samples while reducing that of genuine samples.
In the inference phase, for a given multimodal video, the multimodal classification head is initially utilized to obtain video-level prediction of the modality-specific forgery type, which is exploited for feature selection and localization head selection for temporal forgery localization.
For example, if both modalities are predicted as forged, multimodal features and the corresponding localization head are used for timestamp prediction.
If only the visual modality is forged, enhanced visual features and the visual localization head are employed.
The modality-aware expert selection mechanism leverages the simpler task of classification to facilitate the more challenging task of TFL, improving the flexibility and localization precision of WMMT in WS-MTFL.
The experimental results indicate that this strategy could significantly improve the performance of WS-MTFL.

The main contributions are summarized as follows:
\begin{itemize}
	\item
        The multitask learning is introduced to address WS-MTFL, focusing on balancing multimodal interaction and single-modal forgery traces identification.
	\item
        A temporal property preserving attention mechanism is proposed to perceive the intra- and inter-modality deviation, which is applied to generate enhanced visual features, enhanced audio features, and multimodal features.
  	\item
        An extensible deviation perceiving loss is proposed to explore further temporal information for weakly supervised learning, aiming to enlarge the temporal deviation of forged samples while reducing that of genuine samples.
        \item 
        Extensive experiments have demonstrated the effectiveness of each module of WMMT, achieving comparable results to fully supervised approaches in several evaluation metrics.        
\end{itemize}

As an extension version of our previous work \cite{xu2025multimodaldeviationperceivingframework}, this paper extends the model to a multitask learning paradigm to address fine-grained temporal forgery localization.
Furthermore, we explore a wider variety of deviation measure functions and, importantly, incorporate deviation measure objectives into the analysis.

\section{Related Work}
\subsection{Multimodal Deepfake Detection}
With the gradual progression of Deepfake forensics research, research on multimodal approach utilizing both visual and audio information is becoming increasingly popular \cite{jia2024can,xia2024mmnet,chen2023jointly,nirkin2021deepfake, 10081373}.
The primary issue in multimodal Deepfake detection (MDD) is to identify forgery traces from two distinct embedding spaces.
Chugh et al. \cite{chugh2020not} and McGurk et al. \cite{mcgurk1976hearing} extracted the visual and audio features and compared the discrepancies between the two modalities directly.
To fully facilitate the fusion of multimodal features, Zhou et al. \cite{zhou2021joint} conducted joint audio-visual learning to promote the multimodal interaction.
Meanwhile, Yin et al. \cite{yin2024fine} analyzed the relationships of intra- and inter-modality by the heterogeneous graph and achieved the multimodal fine-grained Deepfake classification target.
To tackle the temporal partial forgery localization challenge \cite{cai2022you}, Zhang et al. \cite{zhang2023ummaformer} proposed to predict forged segments by multimodal feature reconstruction.
Nie et al. \cite{nie2024frade} proposed forgery-aware audio-distilled multimodal learning by capturing high-frequency discriminative features.
However, existing MDD approaches have been conducted primarily under the fully supervised paradigm, and the weakly supervised MDD approach is insufficiently explored.

\subsection{Weakly Supervised learning}
There are three typical types of weakly supervised learning: incomplete supervision, where only a subset of training data is given with annotations; inexact supervision, where the training data are given with only coarse-grained annotations; and inaccurate supervision, where the given annotations are not always accurate \cite{zhou2018brief}.
Weakly supervised learning has achieved numerous progresses in object detection \cite{fu2024cf,su2024consistency} and video understanding \cite{li2024complete}.
In the object detection domain, weakly supervised object localization (WSOL) and weakly supervised object detection (WSOD) are treated as two different tasks \cite{zhang2021weakly}.
WSOL mainly aims at entailing the location of a single object utilizing merely image-level annotations \cite{chen2024adaptive}.
While the goal of WSOD is to detect every possible object with image-level annotations.
In the video understanding domain, weakly supervised temporal action localization (WS-TAL) is proposed to predict the category and start-end timestamps of actions within the visual modality of a given video, training with only video-level action category annotations \cite{wang2023temporal}.
Besides, WS-TFL has been proposed for Deepfake detection.
CoDL \cite{10.1109/TIFS.2025.3533906} and CPL \cite{10887844} attempted to identify temporal partial forged segments based on the visual modality.
The aforementioned weakly supervised approaches mainly concentrate the single-modal tasks, with insufficient systematic research on the multimodal weakly supervised learning paradigm.

\subsection{Multitask learning}
Multitask learning (MTL) is an inductive transfer strategy that improves generalization by integrating the domain information of related tasks \cite{caruana1997multitask}.
Formally, given $m$ learning tasks $\left\{\mathcal{T}_i\right\}_{i=1}^m$ where all the tasks are related, MTL aims to learn the $m$ tasks together to improve the learning of a model for each task $\mathcal{T}_i$ \cite{zhang2021survey}.
MTL has several typical applications in the field of natural language processing (NLP) and emotion recognition.
Multilingual machine learning integrates domain information from multiple languages through MTL, which facilitates the generalization and transferability of NLP models \cite{10.1145/3663363, masumura-etal-2018-multi}, such as sentiment classification in Chinese and English \cite{wang-etal-2018-personalized}.
Besides, more stable and reliable emotion recognition models could be obtained by combining audio-text \cite{chuang2020worse} or visual-audio-text \cite{yu2021learning}.
The performance of multitask models is usually sensitive to the relationships between tasks.
Therefore, the tradeoffs between intra- and inter-task relationships are essential.
Ma et al. \cite{10.1145/3219819.3220007} adapt the Mixture-of-Experts (MoE) structure for MTL to manage this tradeoff.
They employed a gating network to select one or more tasks that fit a specific input sample.
In the context of MDD, MTL could effectively integrate information from visual and audio modalities.
Incorporating MoE into the MTL could further enhance the flexibility by enabling input-adaptive expert selection, making MTL particularly well-suited for handling modality-asymmetric forgery and WS-MTFL.
\section{Proposed Framework}
\subsection{Problem Definition}
The WS-MTFL aims at modality-specific forgery type detection and the timestamp localization of forged segments in multimodal Deepfake video, depending solely on the video-level annotations.
Formally, given a set of videos with video-level annotations $\mathcal{D}=\left\{x_n, y_n\right\}_{n=1}^N$, each video $x_n$ is associated with an annotation $y_n \in \mathbb{R}^{4}$, denoting the modality-specific forgery type.
In the training phase, merely $\mathcal{Y} = \left\{y_n\right\}_{i=n}^N$ are accessible for loss calculation and the model parameter learning.
During the inference phase, the WS-MTFL model should predict the forgery type and locate the forged segments of a given video $x$.

\begin{equation}
\hat{z}=\left\{\hat{y},\left\{\left(s_k, e_k\right)\right\}_{k=1}^K\right\}
\end{equation}

\noindent where $\hat{y} \in \mathbb{R}^{4}$ denotes the predicted modality-specific forgery type, $s_k$ and $e_k$ denote the start and end timestamps of the $k$-th forged segment, and $K$ is the total number of forged segments in $x$.

\subsection{Overview}
\begin{figure*}[htbp]  
    \centering
    \includegraphics[width=\textwidth]{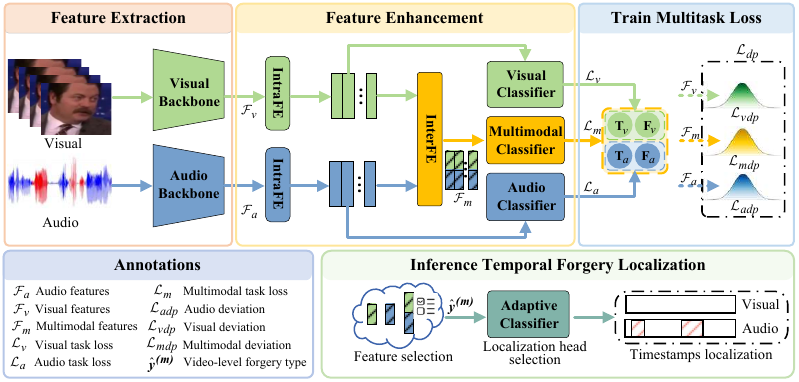}  
    \caption{Overview of the proposed weakly supervised multimodal temporal forgery localization via multitask learning.}
    \label{fig:wmmt_structure}
\end{figure*}

In this paper, a novel weakly supervised multimodal temporal forgery localization via multitask learning (WMMT) is proposed, as shown in Fig. \ref{fig:wmmt_structure}.
The multitask learning (MTL) is introduced to tackle various forgery scenarios.
To reduce the feature interference and improve the localization precision,
the MTL is incorporated with Mixture-of-Experts (MoE) for feature and localization head selection.
A temporal property preserving attention mechanism (TPPA) is proposed for intra-modality feature enhancement (IntraFE) and inter-modality feature enhancement (InterFE).
Additionally, an extensible deviation perceiving loss is proposed to explore further temporal information for WS-MTFL.

Specifically, given an arbitrary video dataset $\mathcal{D}=\left\{x_n, y_n\right\}_{n=1}^N$, the visual features $\mathcal{F}_v$ and audio features $\mathcal{F}_a$ are extracted using pre-trained models for their respective modalities.
The detection for visual and audio modalities is formulated as two tasks, denoted by $\mathcal{T}_v$ and $\mathcal{T}_a$.
For the visual task $\mathcal{T}_v$, IntraFE is applied first to $\mathcal{F}_v$ via the proposed TPPA. 
Subsequently, InterFE is performed to capture inconsistencies between visual and audio modality, resulting in enhanced visual features $\mathcal{F}_v^{\prime\prime}$.
A visual localization head $h_v$ is then trained on $\mathcal{F}_v^{\prime\prime}$ to produce a temporal forgery activation sequence (FAS), denoted by $\mathcal{P}_v = \left\{p_t^{v} \in \mathbb{R}^{2} \right\}_{t=1}^T$, where each $p_t^{v}$ represents the predicted probabilities of the $t$-th visual segment being genuine or forged.
Likewise, we could produce the enhanced audio features $\mathcal{F}_a^{\prime\prime}$ and the audio localization head $h_a$.
To integrate the multimodal information from $\mathcal{T}_v$ and $\mathcal{T}_a$, the enhanced features $\mathcal{F}_v^{\prime\prime}$ and $\mathcal{F}_a^{\prime\prime}$ are concatenated to form multimodal features $\mathcal{F}_m$.
A multimodal localization head $h_m$ is trained on $\mathcal{F}_m$ to produce multimodal FAS, denoted by $\mathcal{P}_m = \left\{p_t^{m} \in \mathbb{R}^{4} \right\}_{t=1}^T$, where each $p_t^{m}$ indicates the predicted modality-specific forgery type of the $t$-th multimodal segment.

In the training phase, three classification heads are trained to predict the video-level modality-specific forgery type for loss calculation and model parameter learning.
For $\mathcal{T}_v$, the visual classification head $c_v$ is to predict the visual modality authenticity $\hat{y}^{(v)} \in \mathbb{R}^{2}$ by aggregating $\mathcal{P}_v$ to train the visual task.
\begin{equation}
\label{frame2video}
\hat{y}_v = \sigma\left(\frac{1}{T}\sum_{t=1}^T p_t^{v}\right)
\end{equation}
where $\sigma(\cdot)$ denotes a normalization operation. 
Similarly, we could obtain audio classification head $c_a$ to predict $\hat{y}^{(a)} \in \mathbb{R}^{2}$, and multimodal classification head $c_m$ to predict $\hat{y}^{(m)} \in \mathbb{R}^{4}$.

During the inference phase, for a given multimodal video, the $c_m$ is initially utilized to predict the video-level modality-specific forgery type $\hat{y}^{(m)}$.
This prediction guides the selection of modality-specific features and localization head to reduce feature interference and improve localization precision (please see Section \ref{sec:MTFLandMTL} for details).
The final prediction consists of the modality-specific forgery type and the corresponding timestamps of forged segments.

\subsection{WS-MTFL under Multitask Learning Paradigm}
\label{sec:MTFLandMTL}
In WMMT, the overall WS-MTFL is formulated as visual task $\mathcal{T}_v$, audio task $\mathcal{T}_a$, and multimodal task $\mathcal{T}_m$, thereby training a model to improve the localization precision under the MTL paradigm.
In this section, we will illustrate how to address the WS-MTFL problem using the MTL paradigm.

\textbf{\textit{1) Single-Modal Modeling}:}
To effectively capture modality-specific forgery traces, we first construct two single-modal binary classification tasks $\mathcal{T}_v$ and $\mathcal{T}_a$.
In the case of $\mathcal{T}_v$, frame-level features of visual modality are first extracted and enhanced, then passed through the visual localization head $h_v$.
The visual classification head $c_v$ produces the visual modality results $\hat{y}_v$ to predict whether the visual modality has been forged.
Due to the lack of fine-grained frame-level annotations, only video-level supervision $y_v \in \mathbb{R}^{2}$ is available during training.
To bridge the gap between coarse video-level supervision and fine-grained temporal localization, we introduce a temporal aggregating mechanism over the frame-level outputs $\mathcal{P}_v = \left\{p_t^{v} \in \mathbb{R}^{2} \right\}_{t=1}^T$ to generate the video-level prediction $\hat{y}_v$, as shown in Eq. (\ref{frame2video}).
The loss $\mathcal{L}_v$ of visual task $\mathcal{T}_v$ is calculated using the binary cross-entropy (BCE) loss:
\begin{equation}
\begin{aligned}
    \mathcal{L}_{v}\!=\!-\frac{1}{N} \sum_{n=1}^{N}\left[y_n^{(v)} \log(\hat{y}_n^{(v)})\!+\!(1\!-\!y_n^{(v)}) \log (1\!-\!\hat{y}_n^{(v)})\right]\\
    \mathcal{L}_{a}\!=\!-\frac{1}{N} \sum_{n=1}^{N}\left[y_n^{(a)} \log (\hat{y}_n^{(a)})\!+\!(1\!-\!y_n^{(a)}) \log (1\!-\!\hat{y}_n^{(a)})\right]
\end{aligned}
\end{equation}
Similarly, the audio task $\mathcal{T}_a$ is trained using the same scheme with the dedicated loss $\mathcal{L}_a$.

Although each task is trained only with video-level
supervision, the localization heads $h_v$ and $h_a$ are indirectly guided to learn temporal forgery traces.
In practice, we observe that the gradients from the classification heads serve as weak supervisory signals, encouraging the localization heads to capture forged segments, consistent with the paradigm of weakly supervised learning \cite{9926133,10880117, wu2025weakly}.

\textbf{\textit{2) Cross-modal Integration}:}
To further mine multimodal forgery traces to improve the localization precision, we integrate the single-modal tasks into a multimodal task $\mathcal{T}_m$, which jointly leverages visual and audio modalities.
This design is grounded in the MTL paradigm, where related tasks could benefit from cross-modal representations, promoting inductive transfer and reducing overfitting \cite{8374898, zhang2021survey, yu2021learning, 10.1145/3219819.3220007}.
The $\mathcal{T}_m$ aims to capture cross-modal inconsistencies that are indiscernible in individual modality.
For example, visual modality often reveals spatial-temporal artifacts such as facial warping or lip desynchronization, while audio provides prosodic and temporal cues that could expose forgery traces \cite{zhou2021joint, yang2023avoid, nie2024frade,yin2024fine,miao2025multi}..
By integrating these complementary signals, $\mathcal{T}_m$ enables more reliable classification of modality-specific forgery type.

To facilitate effective modality interaction, we propose cross-modal feature enhancement in Section \ref{sec: feaEn}.
For each feature segment of visual modality, attention is computed with respect to audio features, generating enhanced visual features $\mathcal{F}_v^{\prime\prime}$ informed by audio context.
The same is applied in reverse to obtain enhanced audio features $\mathcal{F}_a^{\prime\prime}$.
This bi-directional attention preserves temporal alignment and enables selective enhancement of subtle forgery traces.
These enhanced features are concatenated to form multimodal features $\mathcal{F}_m$.
The $\mathcal{F}_m$ are utilized to train two dedicated heads: a multimodal localization $h_m$ for frame-level forgery probability estimation, and a multimodal classification head $c_m$ for video-level modality-specific forgery type prediction (visual only, audio only, or both).
The multimodal classification loss $\mathcal{L}_{m}$ is defined as:
\begin{equation}
\mathcal{L}_{m} = - \frac{1}{N}\sum_{n=1}^{N} \sum_{c=1}^{C} y_{nc}^{(m)} \log(\hat{y}_{nc}^{(m)})
\end{equation}
where $C$ denotes the number of video-level modality-specific forgery types.

The multimodal task $\mathcal{T}_m$ is also supervised at the video-level annotations to train the classification head, while indirectly guiding the localization head to capture multimodal temporal forgery traces.
By optimizing both visual and audio features in a multimodal task, the model leverages cross-modal complementary information to improve the video-level classification accuracy and temporal localization precision.

\textbf{\textit{3) Modality-Aware Expert Selection}:}
While previous components focus on learning single-modal and cross-modal forgery traces under weak supervision, the inference phase still faces the challenge of avoiding modality interference \cite{li2017feature, theng2024feature} and locating the timestamps of forged segments.
To further enhance localization precision, we introduce a modality-aware expert selection mechanism that leverages the output $\hat{y}^{(m)}$ of $c_m$ to adaptively select the appropriate features and the localization head.

Specifically, during inference, we design a gating operation based on the predicted  $\hat{y}^{(m)}$, which estimates whether the forgery type refers to the visual modality only, the audio modality only, or both.
For example, if $\hat{y}^{(m)}$ denotes that both modalities are forged, the multimodal features $\mathcal{F}_m$ and multimodal localization head $h_m$ are selected to predict the timestamps of forged segments.
If $\hat{y}^{(m)}$ indicates that the visual modality is forged while the audio modality is genuine, the $\mathcal{F}_v^{\prime\prime}$ and $h_v$ will be utilized for localization.

This modality-aware expert selection mechanism acts as a lightweight decision gate in the inference phase, allowing the model to focus on the most informative modality features while suppressing the potential inference introduced by the other modality.
By integrating dedicated task predictions into the localization during the inference phase, our framework exhibits properties analogous to the Mixture-of-Experts (MoE) structure, where modality-specific classification adaptively selects the optimal expert for downstream localization.
Moreover, this design naturally complements the MTL architecture, as the multimodal classification head not only serves as a supervisory component but also guides expert selection at inference.
This mechanism leverages the simpler video-level classification prediction to guide the more challenging frame-level timestamp localization of forged segments, leading to improved localization precision, especially in the presence of partial or asymmetric modality forgery.

\subsection{Feature Enhancement}
\label{sec: feaEn}
Given an untrimmed video, IntraFE and InterFE could reveal further forgery traces for multimodal temporal forgery localization.
The IntraFE improves each modality's ability to independently explore temporal forgery traces, such as deviation between frames in the visual modality, or tempo and tone mutations in audio.
The InterFE is to identify inconsistencies between visual and audio modality,  and exploit the complementary information of inter-modality to mine subtle forgery traces.

\textbf{\textit{1) Intra-modality feature enhancement}:}
The visual features $\mathcal{F}_v$ and audio features $\mathcal{F}_a$ are first obtained by pre-trained models, respectively.
The ultimate goal of feature enhancement is to explore forgery traces, enabling the prediction of the forged segments.
Therefore, it is crucial to ensure that the temporal information is not disturbed during feature processing.

In this paper, we propose a novel TPPA for feature enhancement.
In the case of visual modality, the $\mathcal{F}_v \in \mathbb{R}^{T \times d}$ are linearly projected as $\mathbf{Q}$, $\mathbf{K}$, $\mathbf{V}$.
Then, we calculate the relevance between each visual segment based on the projected visual features.
Thus, a relevance matrix could be obtained $\mathcal{R} = \left[r_{ij}\right]^{T \times T}$, where $r_{ij}$ indicates the relevance between the $i$-th visual segment and the $j$-th segment.
For the $2$-D visual features $\mathcal{F}_v$, the row dimension preserves the temporal information of the corresponding video, which is crucial in temporal forgery localization.
Note that if we directly conduct matrix multiplication of $\mathcal{R}$ and $\mathcal{F}_v$, the obtained visual features have already dropped the temporal information.
To preserve the temporal property, the relevance matrix $\mathcal{R}$ is summed by columns to obtained the matrix $\widehat{\mathcal{R}} = \left[\hat{r}_j\right]^{1 \times T}$, where $\hat{r}_j$ represents the relevance between the visual features $\mathcal{F}_v$ and the $j$-th visual segment.
Finally, the intra-modality enhanced visual features $\mathcal{F}_v^{\prime}$ are derived by multiplying each $\hat{r}_j$ with the $j$-th row of the projected visual features $\mathbf{V}$.
Formally,

\begin{equation}
\mathbf{Q}=\mathcal{F}_v \mathbf{W}_{qry}, \quad \mathbf{K}=\mathcal{F}_v \mathbf{W}_{key}, \quad \mathbf{V}=\mathcal{F}_v \mathbf{W}_{vle} \\
\label{eqqkv}
\end{equation}

\begin{equation}
\widehat{\mathcal{R}}_{v} = \sigma\left(\mathbb{S}\left(\frac{\mathbf{Q} \cdot \mathbf{K}^{\mathbb{T}}}{\sqrt{d}}\right)\right)^{\mathbb{T}}
\end{equation}

\begin{equation}
\mathcal{F}_v^{\prime}=(\widehat{\mathcal{R}}_{v})^{\mathbb{T}}\mathbf{V}
\end{equation}

\noindent where $\mathbf{W}_q$, $\mathbf{W}_k$, $\mathbf{W}_v$ are learnable parameters, $\mathbb{S}$ indicates the column summation, $\sigma$ is the normalization operation, and $\mathbb{T}$ indicates the matrix transpose.

Likewise, the intra-modality enhanced audio features $\mathcal{F}_a^{\prime}$ could be calculated.
The details of the TPPA are shown in Algorithm \ref{alg}.

\begin{algorithm}[t]
\SetKwInput{Para}{\textbf{Parameter}}
\caption{The algorithm of the temporal property preserving attention mechanism}
\label{alg}
\KwIn{Visual features $\mathcal{F}_v$, audio features $\mathcal{F}_a$, parameter $d$, learnable parameters $\mathbf{W}_{qry}$, $\mathbf{W}_{key}$, $\mathbf{W}_{vle}$. }
\KwOut{Enhanced features $\mathcal{F}_v^{\prime}$ and $\mathcal{F}_a^{\prime}$.}
Calculate the $\mathbf{Q}$, $\mathbf{K}$, $\mathbf{V}$ in Eq. \ref{eqqkv}\;
Calculate the relevance matrix $\mathcal{R}=\frac{\mathbf{Q} \cdot\mathbf{K}^{\mathbb{T}}}{\sqrt{d}}$\;
\For{$j = 1 : T$}
{
    $\hat{r}_{j} = \sum_{i=1}^{T}\mathcal{r}_{ij}$\;
}
Normalize $\left\{\hat{r}_j \right\}_{t=1}^T$, $\widehat{\mathcal{R}}= \left[\hat{r}_t\right]^{1 \times T}$\;
Calculate the intra-modality enhanced visual features $\mathcal{F}_v^{\prime} = \widehat{\mathcal{R}}^{\mathbb{T}} \cdot \mathbf{V}$\;
Similarly, calculate $\mathcal{F}_a^{\prime}$\;
\textbf{Return} $\mathcal{F}_v^{\prime}$ and $\mathcal{F}_a^{\prime}$
\end{algorithm}

\textbf{\textit{2) Inter-modality feature enhancement}:}
While IntraFE enables each modality to emphasize its internal forgery traces, it is often insufficient for identifying inter-modality inconsistencies.
To address this limitation, we introduce InterFE to enhance each modality feature by leveraging complementary information from the other.

In the case of visual modality, the cross-modal attention of audio features to visual features is also calculated based on Algorithm \ref{alg}.
The relevance between audio and visual features should be calculated in InterFE.
Specifically, we project visual features $\mathcal{F}_v^{\prime} \in \mathbb{R}^{T \times d^{\prime}}$ and audio features $\mathcal{F}_a^{\prime} \in \mathbb{R}^{T \times d^{\prime}}$ into a shared embedding space using learnable parameters.
Then, the relevance matrix $\widehat{\mathcal{R}}_{a2v} = \left[\hat{r}_j^{a2v}\right]^{1 \times T}$ is calculated, where $\hat{r}_j^{a2v}$ denotes the relevance between audio features and the $j$-th visual segment.
Finally, the inter-modality enhanced visual features $\mathcal{F}_v^{\prime\prime}$ are obtained by multiplying each $\hat{r}_j^{a2v}$ with the $j$-th row of projected audio features.
Formally,

\begin{equation}
\mathbf{Q}_v = \mathcal{F}_v^{\prime} \mathbf{W}_{qry}^v, \quad
\mathbf{K}_a = \mathcal{F}_a^{\prime} \mathbf{W}_{key}^a, \quad
\mathbf{V}_a = \mathcal{F}_a^{\prime} \mathbf{W}_{vle}^a
\end{equation}

\begin{equation}
\widehat{\mathcal{R}}_{a2v} = \sigma\left(\mathbb{S}\left(\frac{\mathbf{Q}_v \cdot \mathbf{K}_a^{\mathbb{T}}}{\sqrt{d^{\prime}}}\right)\right)
\end{equation}

\begin{equation}
\mathcal{F}_v^{\prime\prime} = (\widehat{\mathcal{R}}_{a2v})^{\mathbb{T}}  \mathbf{V}_a
\end{equation}

Similarly, the inter-modality enhanced audio features $\mathcal{F}_a^{\prime\prime}$ could be computed by reversing the attention direction.
This mutual enhancement process allows the model to capture cross-modal inconsistencies for further forgery traces.

\subsection{Deviation Perceiving loss}

\begin{figure}[!t]
\centering
\includegraphics[width=\linewidth]{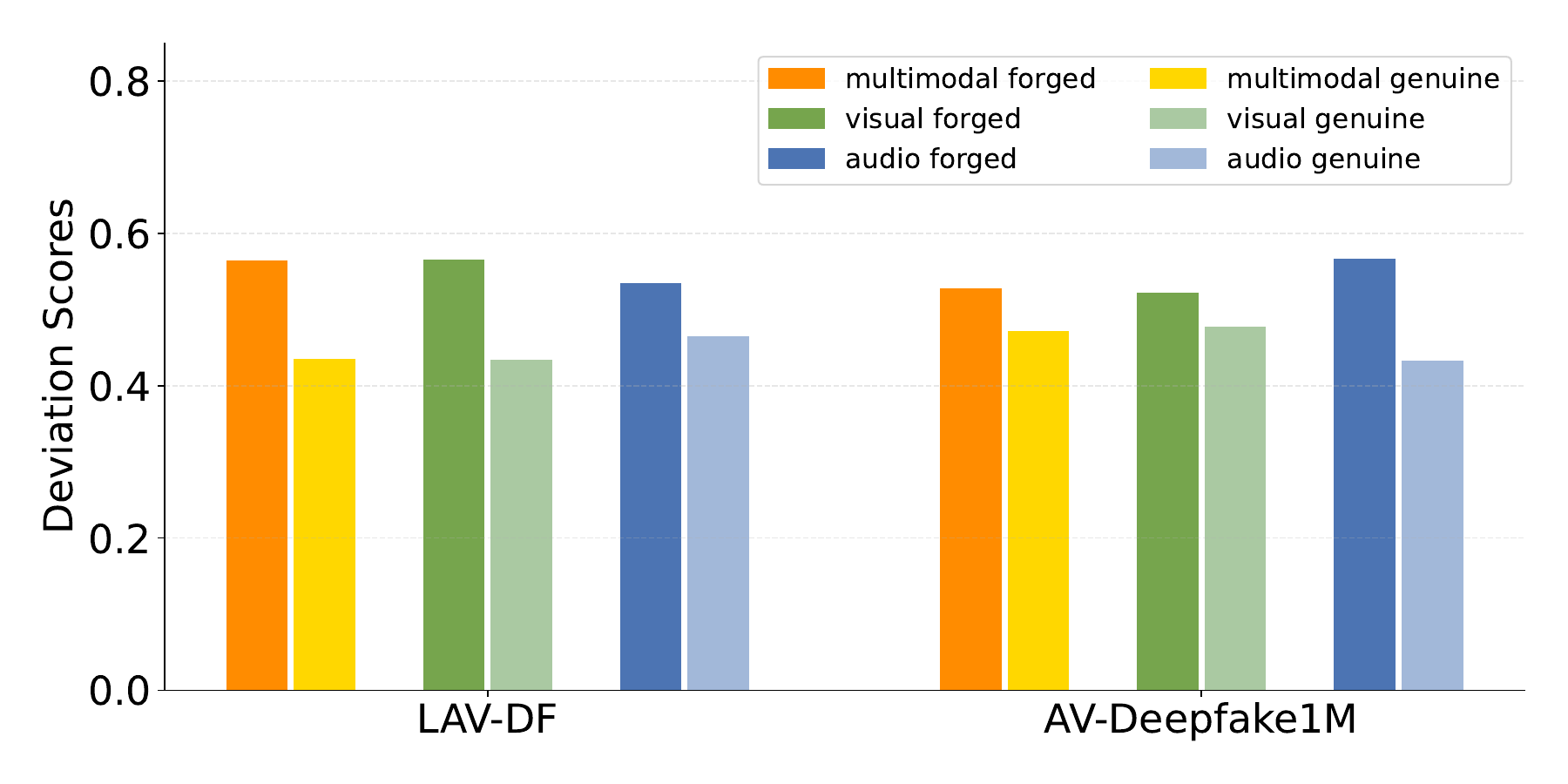}
\caption{Statistical analysis of temporal deviation on the LAV-DF \cite{cai2023glitch} and AV-Deepfake1M \cite{cai2024av} datasets.
The forged visual, audio, and multimodal (visual + audio) samples exhibit greater temporal deviation compared to their corresponding genuine samples.}
\label{fig:deviation}
\end{figure}

The WS-MTFL has merely video-level annotations, which makes it difficult to validly exploit temporal information for the timestamp localization of the forged segments.
Therefore, we need to dig deeper into temporal information for weakly supervised learning.
Typically, video samples obtained from devices like video cameras or smartphones exhibit minimal changes in content and statistical properties between adjacent frames, both in visual and audio modality. In contrast, forged samples are often created by splicing forged frames together, and the forged frames are often obtained by a deep learning model \cite{yin2024fine}.
Maintaining content coherence among the frame-by-frame spliced forged segments is challenging.
As a result, there are often considerable deviations between adjacent forged frames, as well as between these forged frames and the genuine frames.
The statistical analysis of the temporal deviation is shown in Fig. \ref{fig:deviation}.


Considering that for temporal partial forgery samples, the temporal deviation will be larger than that of the genuine samples \cite{zhu2024deepfake,lu2023detection}.
An extensible deviation perceiving loss is proposed to explore further temporal information for weakly supervised learning.
Specifically, in the case of multimodal features $\mathcal{F}_m = \left\{m_t\right\}_{t=1}^{T}$, we calculate the temporal deviation $d$ based on the deviation of adjacent segments.
Formally,
\begin{equation}
    d=\sigma\left(\sum_{t=1}^T f\left(m_t, m_{t+1}\right)\right)
\end{equation}
\begin{equation}
f(m_t, m_{t+1})= E((m_t - m_{t+1})^2)
\label{eq:mse}
\end{equation}
where $f\left(m_t, m_{t+1}\right)$ indicates the deviation between the $t$-th segment and the $(t+1)$-th segment, and $f(\cdot)$ is a deviation measure function that measures the degree of deviation (\textit{e.g.}, mean square error (MSE) as shown in Eq. \eqref{eq:mse}).
The temporal deviation $d$ of forgery samples is commonly larger than that of the genuine samples due to the perturbation of the forged segments.
The multimodal deviation perceiving loss $\mathcal{L}_{mdp}$ is calculated as:
\begin{align}
\mathcal{L}_{mdp} = -\frac{1}{N} \sum_{n=1}^{N} \bigg[ 
    &\ \psi(y_n^{(m)}) \log(d_n) \notag \\
    + &\ (1 - \psi(y_n^{(m)})) \log(1 - d_n) 
\bigg]
\end{align}
where $\psi(y^{(m)})$ is a binary indicator defined as:
\begin{equation}
\psi(y^{(m)})= \begin{cases}1, & \text { if } y^{(m)} \in\{1,2,3\} \\ 0, & \text { if } y^{(m)}=0\end{cases}
\end{equation}
The four labels denote: 0(both genuine), 1(both forged), 2(visual forged only), 3(audio forged only).

Similarly, the objective of deviation perception could be visual or audio features, resulting in $\mathcal{L}_{vdp}$ and $\mathcal{L}_{adp}$, respectively.
In Section \ref{sec:DPA}, we perform a comprehensive analysis of both the function $f(\cdot)$ and the objective of deviation perception, and denote the optimal combination as $\mathcal{L}_{dp}$.

\subsection{Training and Inference}
For WS-MTFL, only video-level annotations are available as supervised signals to train the model.
As illustrated in Section \ref{sec:MTFLandMTL}, the WMMT is formulated as three tasks: visual task $\mathcal{T}_v$, audio task $\mathcal{T}_a$, and multimodal task $\mathcal{T}_m$.
The corresponding losses $\mathcal{L}_v$, $\mathcal{L}_a$, and $\mathcal{L}_m$ for them are calculated relying on the video-level classification error.
Additionally, the deviation perceiving loss is introduced to explore further temporal information for weakly supervised learning.
Therefore, the total loss for training the WMMT is defined as:
\begin{equation}
    \mathcal{L} = \lambda_m\mathcal{L}_{m} + \lambda_v\mathcal{L}_{v} + \lambda_a\mathcal{L}_{a} + \phi\mathcal{L}_{dp}
\end{equation}
where $\lambda_m=0.8$, $\lambda_v = \lambda_a  = \phi =0.1$ are hyperparameters to balance the relationship between different losses.

In the inference phase, given a video $x$, WMMT first predicts the video-level modality-specific forgery type $\hat{y}^{(m)}$ using the multimodal classification head $c_m$.
Then the $\hat{y}^{(m)}$ is utilized to adaptively select appropriate features and localization head for predicting the timestamps of forged segments.
Finally, the WMMT outputs the predicted modality-specific forgery type and corresponding timestamps of forged segments $\hat{z}=\left\{\hat{y}^{(m)},\left\{\left(s_k, e_k\right)\right\}_{k=1}^K\right\}$.

\section{Experiments}
\subsection{Experimental Setup}
\textbf{\textit{1) Datasets}:}
We conduct experiments on two challenging Deepfake temporal partial forgery datasets LAV-DF \cite{cai2023glitch} and AV-Deepfake1M \cite{cai2024av}.
LAV-DF is a strategic content-driven multimodal forgery dataset, which contains \num{36431} genuine videos and \num{99873} forged videos. The duration of forged segments is in the range of $[0-1.6s]$.
AV-Deepfake1M is a large-scale multimodal forgery dataset which contains \num{2068} subjects resulting in \num{286721} genuine videos and \num{860039} forged videos.
There are four types of samples (both genuine, both forged, visual forged only, audio forged only) in both LAV-DF and AV-Deepfake1M.

\textbf{\textit{2) Evaluation metrics}:}
The mean Average Precision (mAP) and average recall (AR) are utilized as the evaluation metrics following \cite{cai2023glitch} and \cite{zhang2023ummaformer}.
For both LAV-DF and AV-Deepfake1M, the IoU thresholds of mAP are set as $[0.1:0.1:0.7]$, and the number of proposals is set as $20$, $10$, $5$, and $2$, respectively.

\textbf{\textit{3) Implementation details}:}
We first extract features from the visual and audio modalities using pre-trained models.
Specifically, we use TSN \cite{wang2018temporal} to extract visual frame features and Wav2Vec \cite{baevski2020wav2vec} to extract audio frame features.
Due to the higher temporal resolution of audio, its frame count differs from that of visual data. To align the temporal dimensions, we apply temporal pooling to audio features, enabling effective multimodal alignment and cross-modal interaction.
The Adam optimizer and cross-entropy loss are used to optimize the proposed WMMT.
The batch size and the learning rate are set to $32$ and $10^{-5}$, respectively.

\begin{table*}[t]
\centering
\caption{Temporal forgery localization results of both fully supervised and weakly supervised approaches on LAV-DF.
For weakly supervised approaches, the best average mAP and AR are presented in \textcolor{red}{\textbf{red}}.
The second-best values are in \textcolor{blue}{\textbf{blue}}.}
\label{lavdf_comp}
\begin{tabularx}{\textwidth}{c|c|XXXXXXXX|XXXXX}
\hline \multirow{2}{*}{ Method } & \multirow{2}{*}{ Supervision } & \multicolumn{8}{c|}{ mAP@IoU(\%) } & \multicolumn{5}{c}{ AR@Proposals(\%) } \\
& & 0.1 & 0.2 & 0.3 & 0.4 & 0.5 & 0.6 & 0.7 & Avg. & 20 & 10 & 5 & 2 & Avg. \\
\hline ActionFormer \cite{zhang2022actionformer} & \multirow{4}{*}{ fully} & 97.97&97.69&97.27&96.78&96.28&95.51&94.61 & 96.59 & 99.17&99.02&98.41&95.95 & 98.14 \\
TriDet \cite{shi2023tridet} &  & 94.99&94.74&94.35&93.83&93.19&92.20&90.67& 93.42 & 97.12&96.93&96.31&93.68 & 96.01 \\
UMMAFormer \cite{zhang2023ummaformer} & & 97.69&97.57&97.37&97.11&96.70&95.96&94.90 & 96.76 & 98.63&98.53&98.22&95.14 & 97.63 \\
MFMS \cite{zhang2024mfms} & & 98.00&97.91&97.78&97.63&97.31&96.69&95.79 & 97.30 & 98.94&98.86&98.62&95.61 & 98.01 \\
\hline CoLA \cite{zhang2021cola} & \multirow{6}{*}{ weakly} & 31.25 & 25.93 & 19.42 & 13.35 & 8.59 & 5.23 & 2.71 &  15.21 & 41.19 & 41.18 & 40.84 & 37.79 & 40.25 \\
FuSTAL \cite{feng2024full} &  &31.55&25.40&19.16&13.48&8.91&5.58&2.95& 15.29 & 39.07&39.05&38.75&36.09 & 38.24 \\
SAL \cite{li2025multilevel} &  & 12.72 & 3.86 & 1.78 & 1.04 & 0.58 & 0.26 & 0.09 & 2.90 & 15.56 & 15.53 & 15.47 & 14.55 & 15.28 \\
LOCO \cite{wu2025weakly} &  & 62.40&55.09&50.78&45.18&36.84&31.65&28.02 & \textcolor{blue}{\textbf{44.28}} & 52.31&52.31&52.31&52.19& \textcolor{blue}{\textbf{52.28}} \\
\cellcolor[rgb]{ .906,  .902,  .902}WMMT & & \cellcolor[rgb]{ .906,  .902,  .902}92.28 & \cellcolor[rgb]{ .906,  .902,  .902}86.67 & \cellcolor[rgb]{ .906,  .902,  .902}80.40 & \cellcolor[rgb]{ .906,  .902,  .902}74.74 & \cellcolor[rgb]{ .906,  .902,  .902}68.23 & \cellcolor[rgb]{ .906,  .902,  .902}60.94 & \cellcolor[rgb]{ .906,  .902,  .902}49.63 & \cellcolor[rgb]{ .906,  .902,  .902}\textcolor{red}{\textbf{73.27}} & \cellcolor[rgb]{ .906,  .902,  .902}85.56 & \cellcolor[rgb]{ .906,  .902,  .902}85.56 & \cellcolor[rgb]{ .906,  .902,  .902}85.56 & \cellcolor[rgb]{ .906,  .902,  .902}85.36 & \cellcolor[rgb]{ .906,  .902,  .902}\textcolor{red}{\textbf{85.51}} \\
\hline
\end{tabularx} 
\end{table*}
\begin{table*}[t]
\centering
\caption{Temporal forgery localization results of both fully supervised and weakly supervised approaches on AV-Deepfake1M.}
\label{av1m_comp}
\begin{tabularx}{\textwidth}{c|c|XXXXXXXX|XXXXX}
\hline \multirow{2}{*}{ Method } & \multirow{2}{*}{ Supervision } & \multicolumn{8}{c|}{ mAP@IoU(\%) } & \multicolumn{5}{c}{ AR@Proposals(\%) } \\
& & 0.1 & 0.2 & 0.3 & 0.4 & 0.5 & 0.6 & 0.7 & Avg. & 20 & 10 & 5 & 2 & Avg. \\
\hline ActionFormer \cite{zhang2022actionformer} & \multirow{4}{*}{ fully} & 67.30&67.27&67.24&67.16&67.02&66.65&65.45 & 66.87&  83.09&82.91&82.56&78.70 & 81.82 \\
TriDet \cite{shi2023tridet} &  & 55.68&55.60&55.43&55.14&54.69&53.82&51.73& 54.58&74.89&74.17&72.85&67.95& 72.47 \\
UMMAFormer \cite{zhang2023ummaformer} & & 91.76&91.67&91.51&91.28&90.97&90.40&88.99 & 90.94 & 95.01&94.59&93.90&89.54 & 93.26 \\
MFMS \cite{zhang2024mfms} & & 94.67&94.63&94.58&94.48&94.32&93.92&92.53 & 94.16 & 96.69&96.43&95.98&91.97 & 95.27 \\
\hline CoLA \cite{zhang2021cola} & \multirow{4}{*}{ weakly} & 3.22&1.09&0.39&0.14&0.05&0.02&0.01 &  \textcolor{blue}{\textbf{0.70}} & 20.71&19.95&16.13&8.17 & \textcolor{blue}{\textbf{16.24}} \\
FuSTAL \cite{feng2024full}  &  & 3.03&1.02&0.40&0.15&0.05&0.02&0.01 & 0.67 &19.67&18.62&14.55&7.08& 14.98 \\
LOCO \cite{wu2025weakly} &  & 1.25&0.30&0.10&0.03&0.01&0.00&0.00 & 0.24 & 10.40&10.11&8.48&3.99& 8.25 \\
\cellcolor[rgb]{ .906,  .902,  .902}WMMT & & \cellcolor[rgb]{ .906,  .902,  .902}54.52 & \cellcolor[rgb]{ .906,  .902,  .902}48.42 & \cellcolor[rgb]{ .906,  .902,  .902}43.55 & \cellcolor[rgb]{ .906,  .902,  .902}38.46 & \cellcolor[rgb]{ .906,  .902,  .902}29.25 & \cellcolor[rgb]{ .906,  .902,  .902}18.8 & \cellcolor[rgb]{ .906,  .902,  .902}7.33 & \cellcolor[rgb]{ .906,  .902,  .902}\textcolor{red}{\textbf{34.33}} & \cellcolor[rgb]{ .906,  .902,  .902}52.83 & \cellcolor[rgb]{ .906,  .902,  .902}52.76 & \cellcolor[rgb]{ .906,  .902,  .902}52.21 & \cellcolor[rgb]{ .906,  .902,  .902}50.17 & \cellcolor[rgb]{ .906,  .902,  .902}\textcolor{red}{\textbf{51.99}} \\
\hline
\end{tabularx} 
\end{table*}
\begin{table*}[t]
\centering
\caption{Generalization performance of cross-dataset evaluation.}
\label{cross_eva}
\begin{tabularx}{\textwidth}{c|c|XXXXXXXX|XXXXX}
\hline \multirow{2}{*}{ Method } & \multirow{2}{*}{ Supervision } & \multicolumn{8}{c|}{ mAP@IoU(\%) } & \multicolumn{5}{c}{ AR@Proposals(\%) } \\
& & 0.1 & 0.2 & 0.3 & 0.4 & 0.5 & 0.6 & 0.7 & Avg. & 20 & 10 & 5 & 2 & Avg. \\
\hline UMMAFormer \cite{zhang2023ummaformer} & \multirow{2}{*}{ fully} & 13.93&13.56&13.13&12.76&12.42&12.14&11.83 & 12.82& 32.35&31.98&31.54&30.43 & 31.58 \\
MFMS  \cite{zhang2024mfms} & & 13.48&12.66&12.01&11.52&11.04&10.59&10.06 & 11.62 &35.38&32.86&30.51&27.49 & 31.56 \\
\hline CoLA \cite{zhang2021cola} & \multirow{3}{*}{ weakly} & 1.35&0.13&0.03&0.01&0.01&0.01&0.01 &  \textcolor{blue}{\textbf{0.22}} & 20.71&19.95&16.13&8.17 & \textcolor{blue}{\textbf{11.06}} \\
LOCO \cite{wu2025weakly} &  & 0.27&0.04&0.01&0.00&0.00&0.00&0.00 & 0.05 & 5.95&5.95&5.72&3.54& 5.29 \\
\cellcolor[rgb]{ .906,  .902,  .902}WMMT & & \cellcolor[rgb]{ .906,  .902,  .902}21.67 & \cellcolor[rgb]{ .906,  .902,  .902}18.50 & \cellcolor[rgb]{ .906,  .902,  .902}15.70 & \cellcolor[rgb]{ .906,  .902,  .902}13.05 & \cellcolor[rgb]{ .906,  .902,  .902}10.21 & \cellcolor[rgb]{ .906,  .902,  .902}7.39 & \cellcolor[rgb]{ .906,  .902,  .902}4.45 & \cellcolor[rgb]{ .906,  .902,  .902}\textcolor{red}{\textbf{13.00}} & \cellcolor[rgb]{ .906,  .902,  .902}63.82 & \cellcolor[rgb]{ .906,  .902,  .902}63.82 & \cellcolor[rgb]{ .906,  .902,  .902}63.81 & \cellcolor[rgb]{ .906,  .902,  .902}60.88 & \cellcolor[rgb]{ .906,  .902,  .902}\textcolor{red}{\textbf{63.08}} \\
\hline
\end{tabularx} 
\end{table*}

\begin{table*}[t]
\centering
\caption{Experimental analysis on the impact of multitask learning for single-modal tasks $\mathcal{T}_a$ and $\mathcal{T}_v$ on LAV-DF.} 
\label{MTL_ana}
\begin{tabularx}{\textwidth}{c|XXXXXXXc|XXXXc}
\hline \multirow{2}{*}{Task}  & \multicolumn{8}{c|}{ mAP@IoU(\%) } & \multicolumn{5}{c}{ AR@Proposals(\%) } \\
&  0.1 &0.2& 0.3 & 0.4 & 0.5 & 0.6 & 0.7 & Avg. & 20 & 10 & 5 & 2 & Avg. \\
\hline
$\mathcal{T}_a$ only& 96.58 & 94.94 & 92.99 & 90.65 & 87.37 & 83.59 & 76.69 & 88.97 & 94.23 & 94.23 & 94.23 & 94.21 & 94.23 \\
$\mathcal{T}_v$ only&87.70 & 79.60 & 68.39 & 56.45 & 43.05 & 31.86 & 19.45 & 55.21 & 73.37 & 73.37 & 73.37 & 72.87 & 73.25 \\
$\mathcal{T}_a$ with MTL&95.94 & 94.25 & 92.30 & 90.07 & 87.44 & 84.72 & 81.35 & 89.44(\textcolor{blue}{\textbf{$\uparrow$0.47}}) & 94.41 & 94.41 & 94.41 & 94.39 & 94.41(\textcolor{blue}{\textbf{$\uparrow$0.18}}) \\
$\mathcal{T}_v$ with MTL& 90.91 & 79.87 & 69.58 & 60.87 & 51.46 & 42.48 & 28.67 & 60.55(\textcolor{red}{\textbf{$\uparrow$5.34}}) & 79.83 & 79.83 & 79.83 & 79.02 & 79.63(\textcolor{red}{\textbf{$\uparrow$6.38}}) \\
 \hline
\end{tabularx}
\end{table*}

\subsection{Intra-Dataset Evaluation}
In this section, we compare the proposed WMMT with previous state-of-the-art approaches on LAV-DF and AV-Deepfake1M.
All comparison approaches were retrained on the pre-trained features according to the open-source code in the paper.

\textit{\textbf{LAV-DF Dataset:}}
As shown in Table \ref{lavdf_comp}, the results show that WMMT achieves relatively good performance on both mAP and AR.
Compared to the WS-TAL approaches \cite{zhang2021cola, feng2024full, li2025multilevel}, the WMMT is significantly improved in both mAP and AR.
It could be found that the AR@$2$ to AR@$20$ remain consistent, which indicates the proposed WMMT could predict the forged segments with fewer candidate proposals.
Compared to the audio WS-TFL approach \cite{wu2025weakly}, WMMT still achieves better localization performance, which demonstrates its advantage in tackling MDD scenarios.
Obviously, the fully supervised approaches \cite{zhang2022actionformer, shi2023tridet, zhang2023ummaformer, zhang2024mfms} achieve superior performance on both mAP and AR.
Such results are also reasonable, as fully supervised approaches are more adept at learning the relationship between the Deepfake video features and corresponding timestamps of forged segments with provided frame-level annotations.
Nevertheless, it could be observed that the WMMT also achieves relatively good performance on mAP@$0.1$, mAP@$0.2$, and AR@$2$ with a relatively small gap with fully supervised approaches.
Overall, the WMMT could effectively identify temporal forgery traces present within the multimodal features, enabling relatively precise localization of timestamps for forged segments.
This primarily benefits from the flexibility of MTL in balancing multimodal interaction and single-modal forgery traces identification. 
Furthermore, temporal deviation perceiving provides WMMT with more temporal information to compensate for the limitations of insufficient annotation.

\textit{\textbf{AV-Deepfake1M Dataset:}}
The average forgery ratio for the AV-Deepfake1M is about $3.6\%$, compared to $7.6\%$ for LAV-DF, presenting an even greater challenge.
The experimental results on AV-DeepFake1M are shown in Table \ref{av1m_comp}.
Experimental results show that fully supervised approaches, ActionFormer and TriDet, exhibit obvious performance degradation because they cannot effectively predict modality-specific forgery type.
A similar trend is observed in all weakly supervised approaches.
Despite the challenging setting, WMMT consistently achieves the best mAP and AR, and obtains results comparable to those of the fully supervised ActionFormer \cite{zhang2022actionformer} and TriDet \cite{shi2023tridet} in mAP@$0.1$.
This advantage can be attributed to the flexibility introduced by WMMT’s multitask learning paradigm, which enables the model to effectively adapt to diverse multimodal forgery scenarios by jointly modeling modality-specific and cross-modal representations.

\subsection{Cross-Dataset Evaluation}
To evaluate the generalization capability of the proposed WMMT, we conduct a cross-dataset evaluation in this section.
Specifically, the model is trained on the AV-DeepFake1M dataset and evaluated on the LAV-DF dataset without any fine-tuning.
As shown in Table \ref{cross_eva}, all approaches experience a noticeable performance drop.
For fully supervised approaches, the performance degradation can be largely attributed to their limited capacity in modeling modality-specific forgery type.
For example, UMMAFormer \cite{zhang2023ummaformer} and MFMS \cite{zhang2024mfms} combine visual and audio features through direct concatenation, lacking specialized designs for multimodal fine-grained Deepfake detection.
In the case of weakly supervised approaches, the primary cause of performance degradation is the deficiency in predicting the timestamps of forged segments.
In contrast, WMMT achieves relatively good performance in cross-dataset evaluation, particularly in the metric AR.
This is attributed to WMMT's flexibility in handling multimodal features, properly balancing multimodal interaction and single-modal forgery traces identification.
Additionally, the deviation perceiving idea helps the WMMT identify temporal anomalies in forged samples, thereby improving the recall rate of temporal forgery localization.
\subsection{Multitask Learning Analysis}
\label{sec:MLA}
As discussed in Section~\ref{sec:MTFLandMTL}, MTL is employed to jointly optimize several related tasks with the goal of improving generalization and performance on each individual task. To further validate the effectiveness of MTL in addressing multimodal Deepfake localization under weak supervision, we conduct a series of controlled experiments in this section.

Specifically, we design four experimental settings:
(1) training with the audio task only ($\mathcal{T}_a$ only);
(2) training with the visual task only ($\mathcal{T}_v$ only);
(3) jointly training $\mathcal{T}_a$ and $\mathcal{T}_v$ under the MTL paradigm and evaluating on $\mathcal{T}_a$ ($\mathcal{T}_a$ with MTL);
(4) jointly training $\mathcal{T}_a$ and $\mathcal{T}_v$ and evaluating on $\mathcal{T}_v$ ($\mathcal{T}_v$ with MTL).
Since the multimodal task $\mathcal{T}_m$ inherently integrates $\mathcal{T}_a$ and $\mathcal{T}_v$ and is essential to the WMMT framework, its MTL effect is not analyzed separately.
As shown in Table ~\ref{MTL_ana}, jointly optimizing the visual and audio tasks leads to consistent performance improvements for both $\mathcal{T}_v$ and $\mathcal{T}_a$.
The visual task benefits more significantly, achieving an average improvement of mAP (+5.34$\%$) and AR (+6.38$\%$), indicating that complementary audio information effectively enhances visual temporal localization.
Since the audio features contain less redundant information, the localization performance of $\mathcal{T}_a$ is better than $\mathcal{T}_v$.
Nevertheless, MTL further boosts the performance of $\mathcal{T}_a$, highlighting the mutual benefit of joint optimization.
These results validate that MTL facilitates knowledge sharing between modalities, enabling the model to exploit cross-modal cues for better temporal forgery localization.
Furthermore, MTL serves as an effective strategy to balance multimodal interaction and single-modal forgery traces identification within the proposed WMMT framework.

\begin{table*}[t]
\centering
\caption{Experimental comparison of deviation measure \textbf{functions} $f(\cdot)$ for temporal forgery localization on LAV-DF.} 
\label{DMF}
\begin{tabularx}{0.9\textwidth}{c|XXXXXXXX|XXXXX}
\hline \multirow{2}{*}{$f(\cdot)$}  & \multicolumn{8}{c|}{ mAP@IoU(\%) } & \multicolumn{5}{c}{ AR@Proposals(\%) } \\
&  0.1 &0.2& 0.3 & 0.4 & 0.5 & 0.6 & 0.7 & Avg. & 20 & 10 & 5 & 2 & Avg. \\
\hline
$L_1$& 84.17&74.47&66.82&60.82&55.68&51.99&47.83& 63.11 &71.55&71.55&71.55&71.30 & 71.49 \\
$L_2$& 89.45&80.80&72.46&65.15&58.51&54.03&49.15& \textcolor{blue}{\textbf{67.08}} & 76.35&76.35&76.34&76.11 & 76.29\\
$cosine$& 88.46 & 82.81 & 73.84 & 66.12 & 57.67 & 49.04 & 36.31 & 64.89 & 83.38 & 83.38 & 83.38 & 82.69 & \textcolor{blue}{\textbf{83.21}}\\
$D_{KL}$&87.72&80.11&70.69&62.38&55.29&49.99&43.35& 64.22 &77.89&77.89&77.89&77.48& 77.79\\
$MSE$& 92.28 & 86.67 & 80.40 & 74.74 & 68.23 & 60.94 & 49.63 & \textcolor{red}{\textbf{74.91}} & 85.56 & 85.56 & 85.56 & 85.36 & \textcolor{red}{\textbf{85.51}} \\
 \hline
\end{tabularx}
\end{table*}

\begin{table*}[t]
\centering
\caption{Experimental comparison of deviation measure \textbf{objectives} for temporal forgery localization on LAV-DF.} 
\label{DP_ob}
\begin{tabularx}{0.9\textwidth}{c|XXXXXXXX|XXXXX}
\hline \multirow{2}{*}{Objective}  & \multicolumn{8}{c|}{ mAP@IoU(\%) } & \multicolumn{5}{c}{ AR@Proposals(\%) } \\
&  0.1 &0.2& 0.3 & 0.4 & 0.5 & 0.6 & 0.7 & Avg. & 20 & 10 & 5 & 2 & Avg. \\
\hline
$\mathcal{T}_a$ & 87.44 & 78.79 & 69.74 & 62.07 & 55.31 & 50.32 & 44.17 & 61.56 & 74.88 & 74.88 & 74.87 & 74.52 & 74.79 \\
$\mathcal{T}_v$ & 90.71 & 82.96 & 73.57 & 64.84 & 57.12 & 50.77 & 43.45 & 68.25 & 78.30 & 78.30 & 78.30 & 77.98 & 78.22 \\
$\mathcal{T}_m$ & 92.28 & 86.67 & 80.40 & 74.74 & 68.23 & 60.94 & 49.63 & \textcolor{blue}{\textbf{74.91}}& 85.56 & 85.56 & 85.56 & 85.36 & \textcolor{red}{\textbf{85.51}} \\
$\mathcal{T}_a + \mathcal{T}_v$& 90.85 & 86.10 & 78.54 & 71.06 & 63.21 & 56.65 & 49.37 & 73.53 & 81.35 & 81.35 & 81.34 & 80.96 & 81.25 \\
$\mathcal{T}_a + \mathcal{T}_v + \mathcal{T}_m$& 90.99 & 86.69 & 81.17 & 75.75 & 68.83 & 61.29 & 50.82 & \textcolor{red}{\textbf{75.32}} & 84.79 & 84.79 & 84.79 & 84.48 & \textcolor{blue}{\textbf{84.71}} \\
 \hline
\end{tabularx}
\end{table*}

\subsection{Deviation Perceiving Analysis}
\label{sec:DPA}
In this section, we conduct experimental analysis on deviation measure function and deviation measure objective, which aims to explore the details of deviation perceiving loss to mine more discriminative temporal representations for weakly supervised learning.

\textit{\textbf{Deviation Measure Function:}}
While calculating the $\mathcal{L}_{dp}$, a deviation measure function $f(\cdot)$ is required to measure the deviation between adjacent segments.
Several experiments are conducted to test the effectiveness of different $f(\cdot)$ on the performance of temporal forgery localization.
Considering the computational complexity and parallelism, five deviation measure functions, Manhattan distance ($L_{1}$), Euclidean distance ($L_{2}$), Cosine Similarity (cosine), Kullback-Leibler Divergence ($D_{KL}$) and Mean Square Error ($MSE$), are selected for the experiments.
The experimental results are shown in Table \ref{DMF}.
It could be observed that the selection of $f(\cdot)$ has an obvious influence on the performance of WMMT.
$MSE$ achieves the best performance among the five deviation measure functions.
This further illustrates the value of the deviation perceiving idea proposed in MDP for WS-MTFL task.
It should be noted that the deviation measure function discussed in this paper remains an open problem.
Investigating more effective $f(\cdot)$ represents a meaningful research direction.

\textit{\textbf{Deviation Measure Objective:}}
The proposed WMMT consists of three tasks: audio task $\mathcal{T}_a$, visual task $\mathcal{T}_v$, and multimodal task $\mathcal{T}_m$.
Therefore, determining the objective of deviation perception is also an important issue.
We divide the objectives into five types: $\mathcal{T}_a$, $\mathcal{T}_v$, $\mathcal{T}_m$, $\mathcal{T}_a + \mathcal{T}_v$, and $\mathcal{T}_a + \mathcal{T}_v + \mathcal{T}_m$.
AS shown in Table \ref{DP_ob}, the experimental results reveal that $\mathcal{T}_m$ and $\mathcal{T}_a + \mathcal{T}_v + \mathcal{T}_m$ leads to relative better performance.
These results could be attributed to the complementary characteristics of temporal inconsistency in visual and audio modalities.
When both modalities are simultaneously considered in the deviation measure, the WMMT is able to perceive a broader range of temporal artifacts, thereby enhancing the overall localization precision.
The deviation perceiving loss serves as an auxiliary role that guides the WMMT toward more discriminative temporal representations.
Moreover, the joint perceiving of multimodal deviation enables stronger regularization of the latent feature dynamics, reinforcing consistency within genuine samples and amplifying the response to forged samples in either modality.
\begin{table*}[t]
\centering
\caption{Ablation study about IntraFE, MTL, InterFE and $\mathcal{L}_{dp}$ on LAV-DF.}
\label{ablation}
\begin{tabularx}{\textwidth}{cccc|XXXXXXXX|XXXXX}
\hline \multirow{2}{*}{IntraFE} & \multirow{2}{*}{MTL} & \multirow{2}{*}{InterFE}&\multirow{2}{*}{$\mathcal{L}_{dp}$}  & \multicolumn{8}{c|}{ mAP@IoU(\%) } & \multicolumn{5}{c}{ AR@Proposals(\%) } \\
& & & & 0.1 &0.2& 0.3 & 0.4 & 0.5 & 0.6 & 0.7 & Avg. & 20 & 10 & 5 & 2 & Avg. \\
\hline
   & & & & 62.20 & 56.64 & 49.03 & 40.77 & 30.89 & 19.39 & 10.47 & 38.48 & 63.12 & 61.01 & 57.9 & 52.25 & 58.57 \\
$\checkmark$ & & & & 61.83 & 56.58 & 49.61 & 42.75 & 33.14 & 19.8 & 9.31 & 39.00 & 64.79 & 64.62 & 63.84 & 59.42 & 63.17 \\
$\checkmark$ & $\checkmark$ & &  & 90.07 & 84.91 & 77.51 & 69.41 & 60.52 & 53.5 & 46.96 & 68.98 & 80.74 & 80.74 & 80.71 & 80.11 & 80.58 \\
 $\checkmark$ & $\checkmark$ & $\checkmark$ & & 88.85 & 84.27 & 76.97 & 69.87 & 62.95 & 56.34 & 48.16 & \textcolor{blue}{\textbf{69.63}} & 83.92 & 83.92 & 83.91 & 83.44 & \textcolor{blue}{\textbf{83.80}} \\
  $\checkmark$ & $\checkmark$ & $\checkmark$ & $\checkmark$ & 92.28 & 86.67 & 80.4 & 74.74 & 68.23 & 60.94 & 49.63 & \textcolor{red}{\textbf{73.27}} & 85.56 & 85.56 & 85.56 & 85.36 & \textcolor{red}{\textbf{85.51}} \\
 \hline
\end{tabularx}
\end{table*}
\begin{figure*}[htbp]
\captionsetup[subfloat]{font=footnotesize}
  \centering
  \subfloat[audio-Raw]{
    \includegraphics[width=0.28\textwidth]{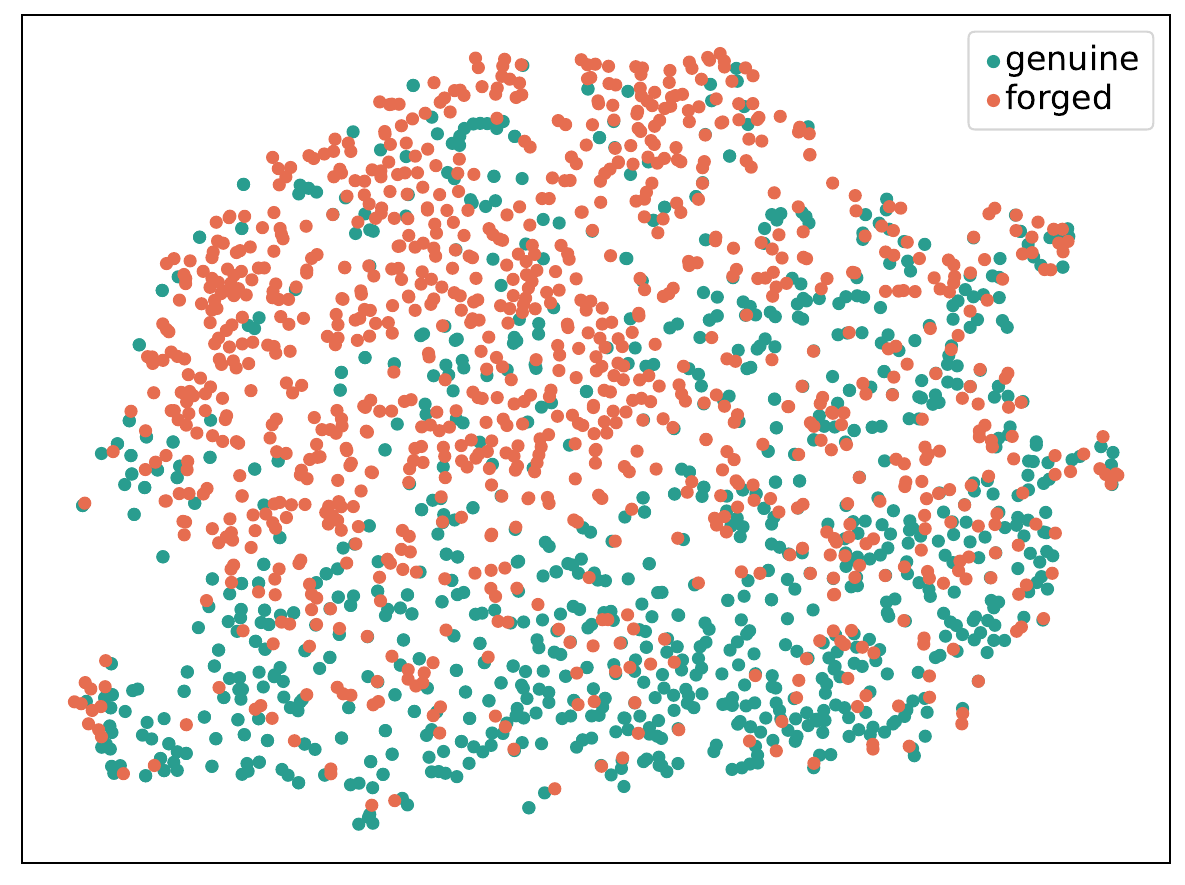}
  }
  \hspace{-0.02\textwidth}
  \subfloat[visual-Raw]{
    \includegraphics[width=0.28\textwidth]{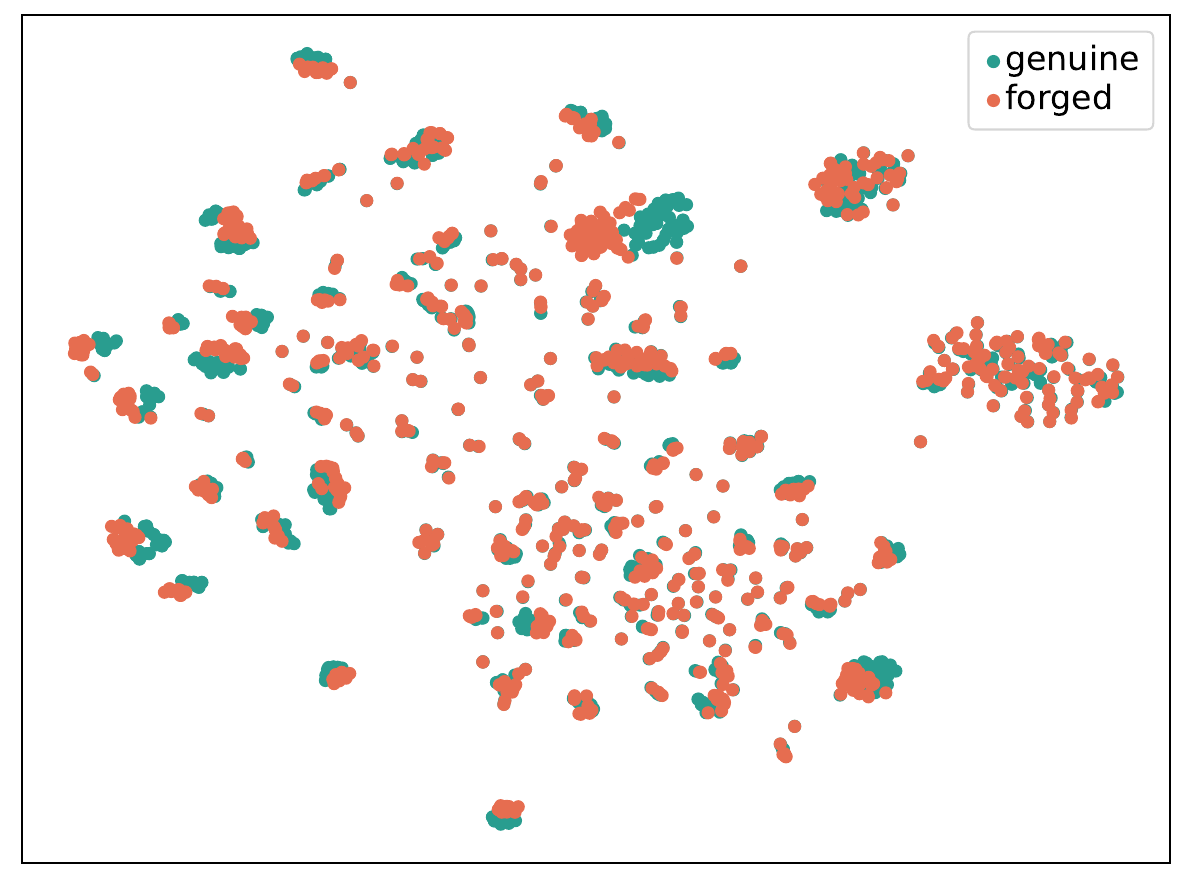}
  }
  \hspace{-0.02\textwidth}
  \subfloat[multimodal-Raw]{
    \includegraphics[width=0.28\textwidth]{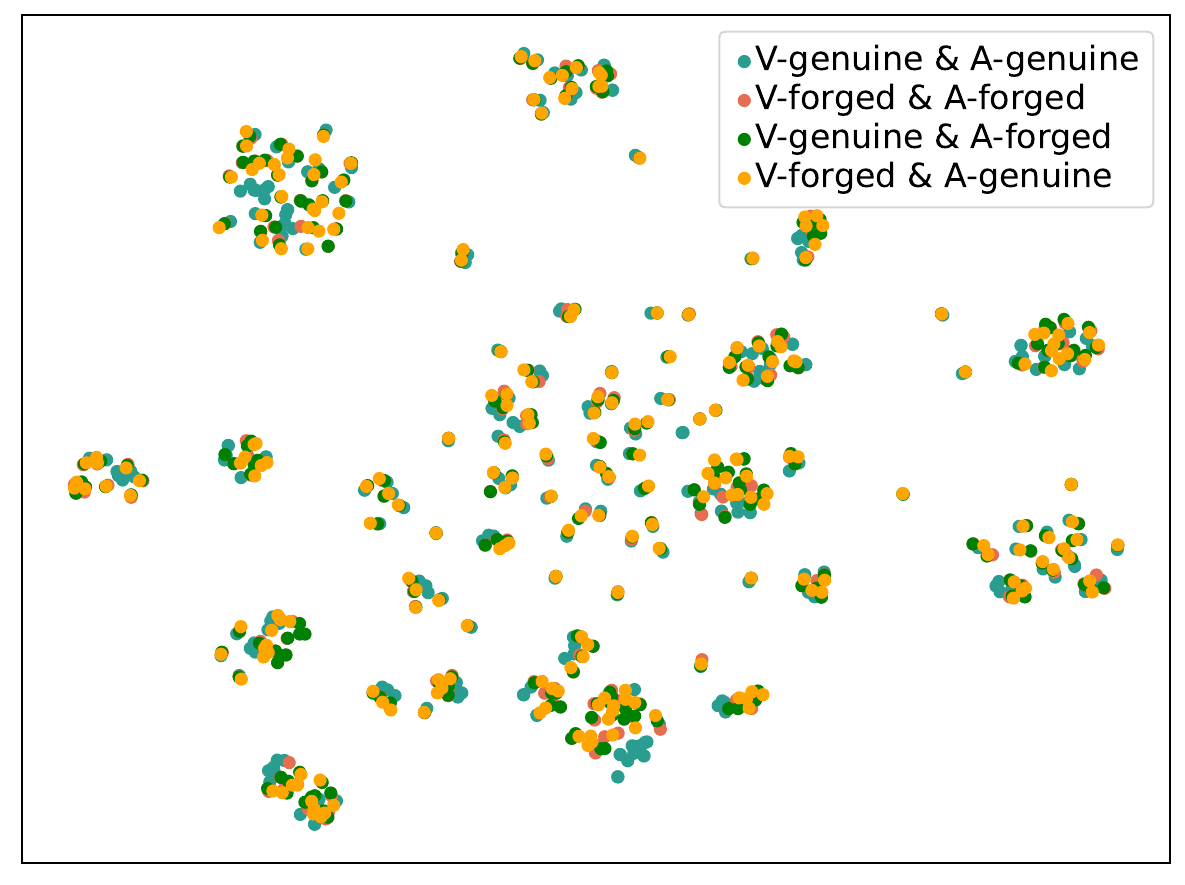}
  }

  \vspace{-1em}  

  \subfloat[audio-WMMT]{
    \includegraphics[width=0.28\textwidth]{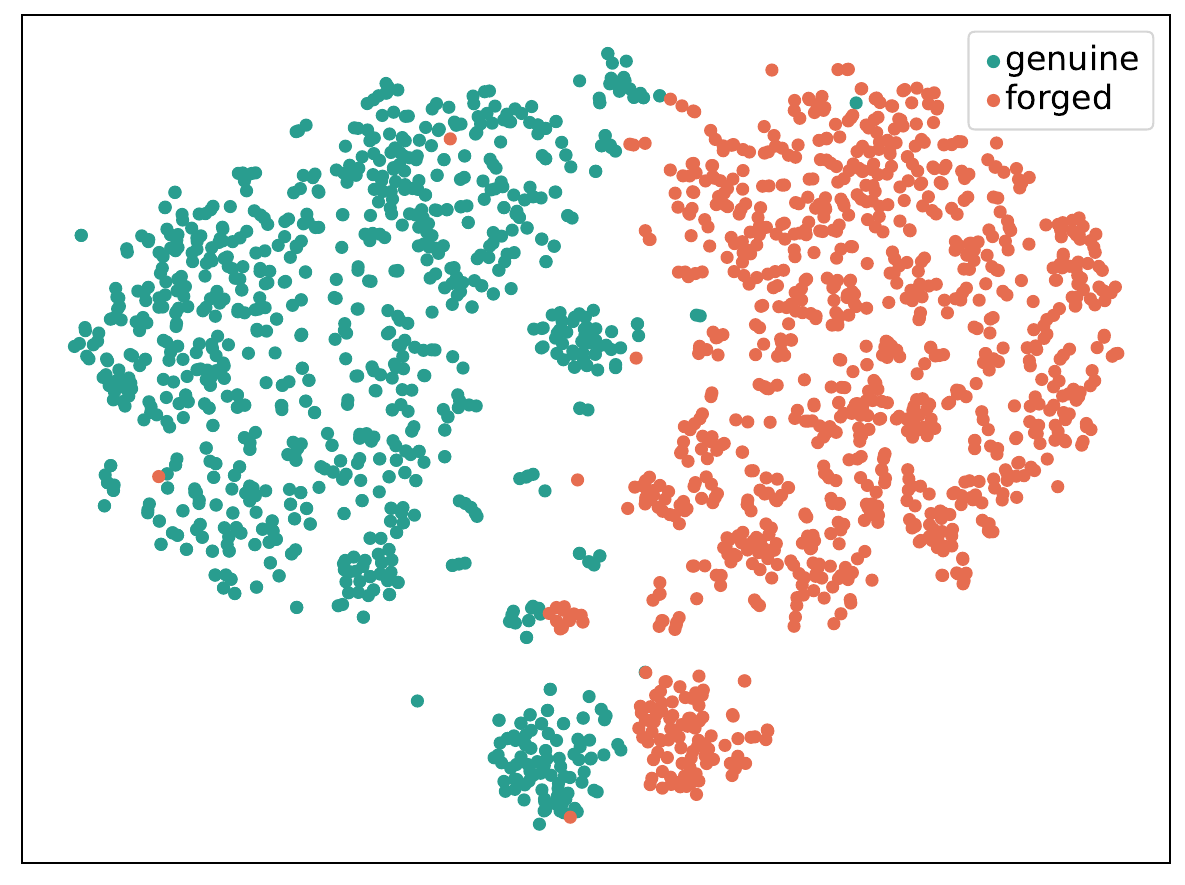}
  }
  \hspace{-0.02\textwidth}
  \subfloat[visual-WMMT]{
    \includegraphics[width=0.28\textwidth]{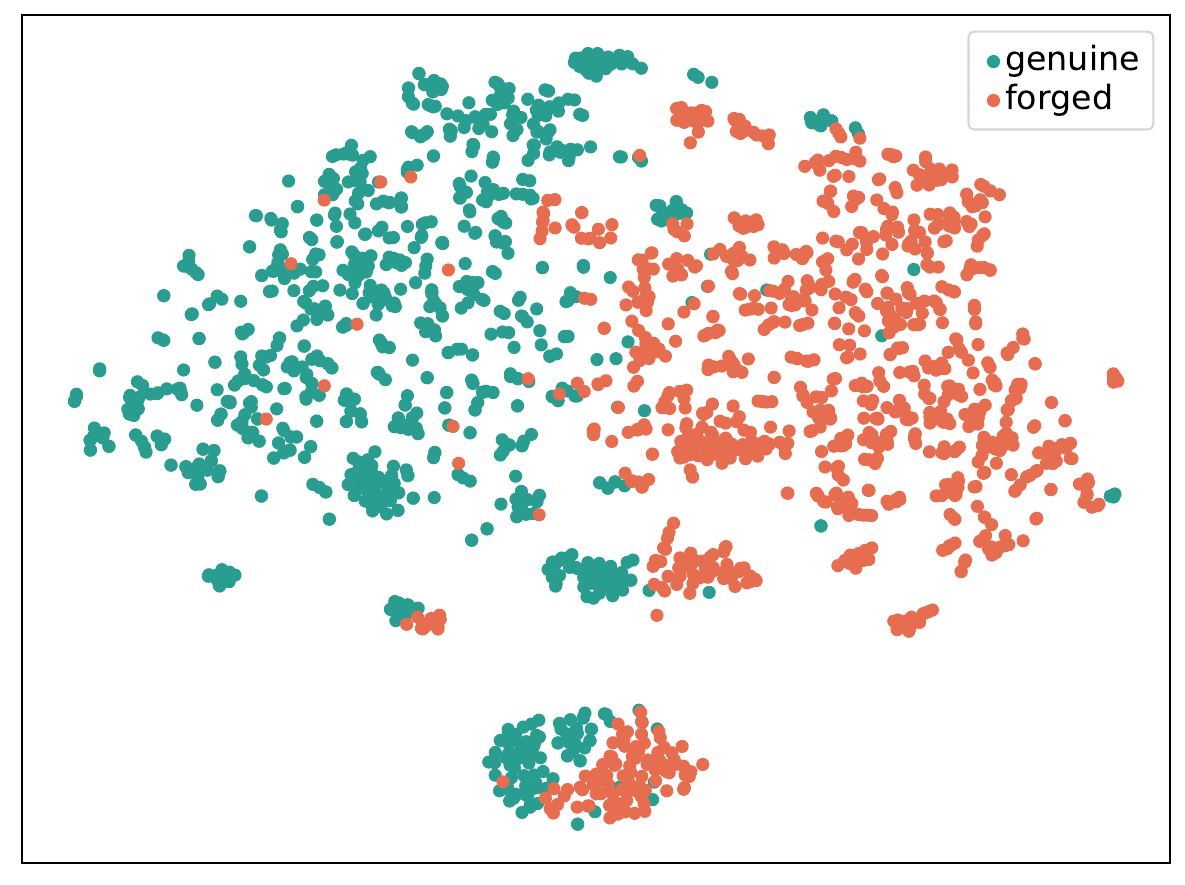}
  }
  \hspace{-0.02\textwidth}
  \subfloat[multimodal-WMMT]{
    \includegraphics[width=0.28\textwidth]{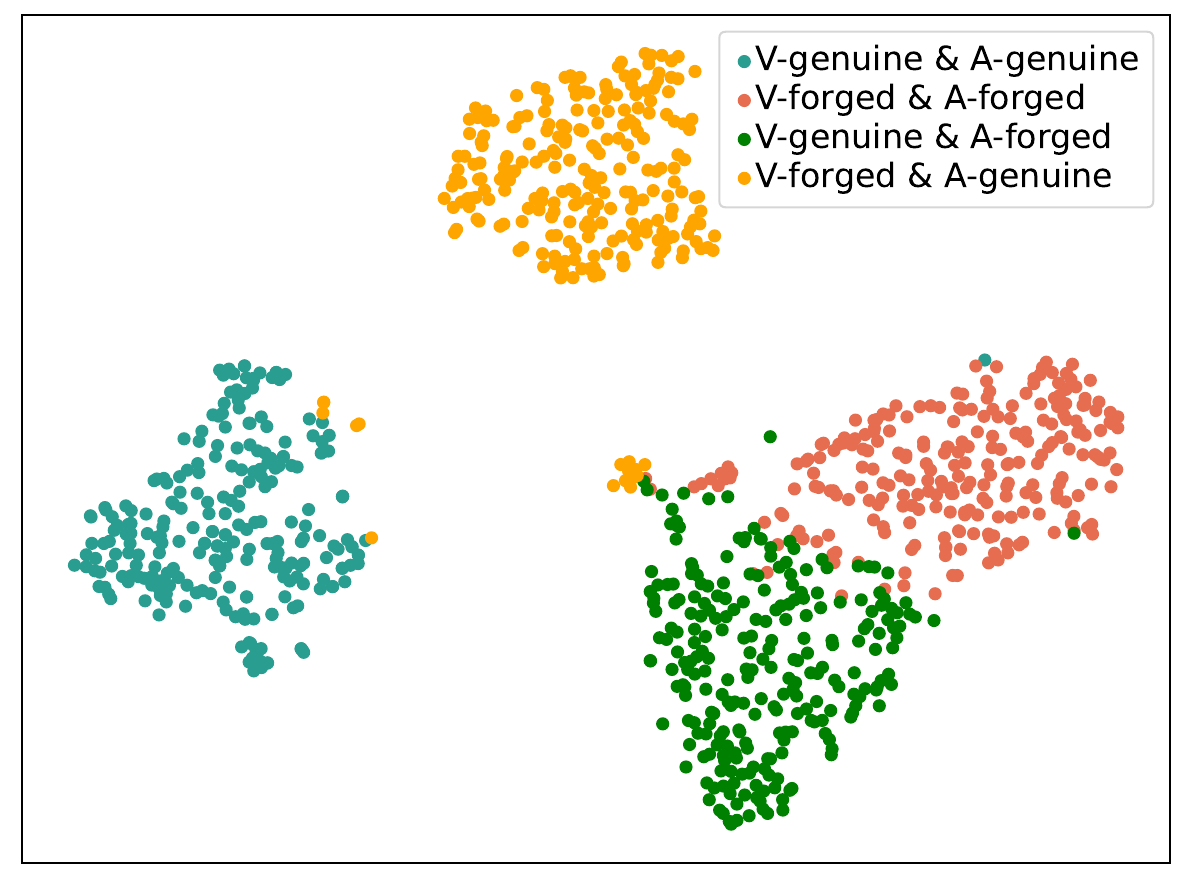}
  }
  \caption{t-SNE visualizations of the video-level representations before and after WMMT enhancement.
(a)-(c) show the pre-trained audio, visual, and multimodal features, while (d)-(f) represent the enhanced features processed by WMMT.}
  \label{fig:tsne}
\end{figure*}
\subsection{Ablation Study}
\label{sec:ablation}
In this section, comprehensive ablation studies are conducted to explore the effectiveness of the three core components in WMMT: multitask learning (MTL), feature enhancement (IntraFE and InterFE), and deviation perceiving loss ($\mathcal{L}_{dp}$).
Specifically, we conducted five experiments.
The baseline is to generate the FAS $\mathcal{P}_m$ and video-level prediction result $\hat{y}^{(m)}$ by directly concatenating visual and audio features together.
The other four experiments verify the temporal forgery performance after sequentially introducing IntraFE, MTL, InterFE, and $\mathcal{L}_{dp}$.

It could be observed that localization performance has gradually improved with the introduction of each component.
In particular, with the assistance of MTL, both average mAP (+29.98$\%$) and AR (+17.41$\%$) have significantly improved.
This demonstrates that MTL improves both the flexibility and localization precision of WMMT, offering a clear advantage for the WS-MTFL task.
The IntraFE and InterFE constructs more discriminative representations by mining intra-modality forgery traces (e.g., adjacent frame deviation in visual modality, or tempo and tone mutations in audio.) and inter-modality inconsistencies, further improving the performance of temporal forgery localization.
In addition, the introduction of deviation perceiving loss $\mathcal{L}_{dp}$ also improves the average mAP (+3.64$\%$) and AR (+1.71$\%$).
The statistical analysis in Fig. \ref{fig:deviation} has shown that forged samples tend to have a larger deviation between adjacent segments compared to genuine samples.
$\mathcal{L}_{dp}$ constrains the WMMT to enlarge the temporal deviation of forgery samples while reducing that of genuine samples, which explores further discriminative temporal information despite the absence of frame-level annotations.
According to Table \ref{ablation}, the best average mAP and average AR are achieved by introducing MTL, FM, and $\mathcal{L}_{dp}$.
It validates the effectiveness of the key components in WMMT.
\subsection{Visualization Analysis}
\label{visual_ana}
To qualitatively evaluate the representation capability of the proposed WMMT, we perform t-SNE visualizations of the learned features across audio, visual, and multimodal modalities.
The visualization consists of three groups of comparisons, each illustrating the separability of modality features before and after feature enhancement.

As shown in Fig. \ref{fig:tsne} (a) and (d), we visualize the raw audio extracted by the pretrained model, alongside the enhanced audio features produced by IntraFE and InterFE.
It can be observed that the enhanced audio features exhibit more distinct clusters, where forged and genuine samples are better separated, demonstrating the WMMT's ability to capture discriminative audio forgery traces.
Fig. \ref{fig:tsne} (b) and (e) depict the same comparison for visual modality features.
Consistent with the audio case, the enhanced visual features exhibit improved cluster separation, suggesting that the visual task of WMMT is capable of amplifying subtle spatial-temporal forgery traces.
Fig. \ref{fig:tsne} (c) and (f) visualize the multimodal features.
We perform fine-grained classification of multimodal features into modality-specific forgery types.
The multimodal features extracted by pre-trained models are difficult to distinguish.
In contrast, the multimodal features enhanced by WMMT exhibit more pronounced separation between different modality-specific forgery types, validating the effectiveness of WMMT in cross-modal interaction and joint modeling.

Overall, the visualizations provide further qualitative evidence that WMMT exhibits reliable discriminability of both single-modal and multimodal representations based on multitask learning and feature enhancement.

\section{Conclusion}
In this paper, we propose a novel weakly supervised multimodal temporal forgery localization via multitask learning (WMMT).
WMMT formulates the WS-MTFL into audio task, visual task, and multimodal task, aiming to balance multimodal interaction and single-modal forgery traces identification.
A temporal property preserving attention mechanism is proposed for intra- and inter-modality feature enhancement to mine subtle forgery traces.
In addition, we introduce an extensible deviation perceiving loss to explore further temporal information for weakly supervised learning.
Extensive experiments demonstrate that WMMT achieves comparable results to fully supervised approaches in several evaluation metrics.
In the future, we will further explore strategies to improve model generalizability.
\bibliographystyle{IEEEtran}
\bibliography{reference}



\end{document}